\documentclass[conference]{IEEEtran}
\IEEEoverridecommandlockouts
\usepackage{cite}
\usepackage{amsmath,amssymb,amsfonts}
\usepackage{algorithmic}
\usepackage{graphicx}
\usepackage{textcomp}
\usepackage{xcolor}
\usepackage[caption=false,font=footnotesize]{subfig} 
\usepackage{siunitx}      
\usepackage[version=4]{mhchem} 
\usepackage{url}          
\usepackage[utf8]{inputenc}
\begin{document}

\title{AQUAIR: A High-Resolution Indoor Environmental Quality Dataset for Smart Aquaculture Monitoring}

\author{
  \IEEEauthorblockN{
    Youssef Sabiri\IEEEauthorrefmark{1}, 
    Walid Houmaidi,
    Ouail El Maadi,
    Yousra Chtouki
  }
  \IEEEauthorblockA{
    School of Science and Engineering, Al Akhawayn University\\
    Ifrane, Morocco\\
    Email: \{y.sabiri, w.houmaidi, o.elmaadi, y.chtouki\}@aui.ma
  }
  \thanks{* Corresponding Author.}
}

\maketitle

\begin{abstract}
Smart aquaculture systems depend on rich environmental data streams to
protect fish welfare, optimize feeding, and reduce energy use.  Yet
public datasets that describe the \emph{air} surrounding indoor tanks
remain scarce, limiting the development of forecasting and
anomaly-detection tools that couple head-space conditions with
water-quality dynamics.  We therefore introduce \textbf{AQUAIR}, an
open‐access public dataset that logs six Indoor Environmental
Quality (IEQ) variables—air temperature, relative humidity, carbon
dioxide, total volatile organic compounds, PM\textsubscript{2.5} and
PM\textsubscript{10}—inside a fish aquaculture facility in Amghass,
Azrou, Morocco.  A single Awair HOME monitor sampled every five minutes
from 14 October 2024 to 9 January 2025, producing more than \mbox{23\,000}
time-stamped observations that are fully quality-controlled and publicly archived on
Figshare.  We
describe the sensor placement, ISO-compliant mounting height, calibration checks
against reference instruments, and an open-source processing pipeline
that normalizes timestamps, interpolates short gaps, and exports
analysis-ready tables.  Exploratory statistics show stable conditions
(median \ce{CO2} = 758 ppm; PM\textsubscript{2.5} = 12\,\si{\micro\gram\per\cubic\metre})
with pronounced feeding-time peaks, offering rich structure for
short-horizon forecasting, event detection, and sensor drift studies.
AQUAIR thus fills a critical gap in smart aquaculture informatics and provides
a reproducible benchmark for data-centric machine learning curricula and
environmental sensing research focused on head-space dynamics in
recirculating aquaculture systems.
\end{abstract}


\section{Introduction}
The global decline of fish populations has become a critical environmental and food security issue. Recent studies highlight that a significant proportion of marine fish species are currently threatened with extinction, reflecting widespread biodiversity loss driven by habitat degradation and climate change \cite{Finn2023MoreLosers}. Trout species, such as the widely farmed rainbow trout (\emph{Oncorhynchus mykiss}), exemplify this vulnerability. These species are not only ecologically important but also economically valuable in aquaculture \cite{russian}, yet they face increasing pressures from disease outbreaks that threaten their survival and productivity \cite{Esposito2024Changes}.

Aquaculture—the controlled cultivation of aquatic organisms including fish, crustaceans, and plants—has emerged as a vital sector to meet the growing global demand for seafood while alleviating pressure on wild fish stocks \cite{Lal2024RobotassistedAA}. It encompasses a variety of systems, from traditional pond culture, which remains widely used due to its simplicity and cost-effectiveness, to advanced recirculating aquaculture systems (RAS), which provide opportunities to reduce water usage and to improve waste management and nutrient recycling \cite{MARTINS201083}. This control is particularly important for sensitive species like trout, which require stable environmental conditions for optimal growth and health \cite{Farimani2025InvestigationOI}. Globally, aquaculture contributes significantly to food security and economic development, especially in regions where natural fisheries are declining or inaccessible.

Indoor Environmental Quality (IEQ) is a key factor influencing aquaculture success, especially in closed or semi-closed facilities where environmental parameters can be closely monitored and managed \cite{article}. IEQ broadly refers to the quality of the indoor environment, including temperature, humidity, carbon dioxide (CO$_2$), volatile organic compounds (VOCs), and particulate matter (PM$_{2.5}$ and PM$_{10}$). For trout, which are particularly sensitive to environmental fluctuations, maintaining optimal IEQ is essential to reduce disease risks and enhance production efficiency \cite{limnolrev24040035}.

Recent advances in machine learning and deep learning have revolutionized environmental monitoring and aquaculture management \cite{Rather2024ExploringAI}. These technologies enable continuous, real-time analysis of complex datasets, facilitating early detection of environmental anomalies and predictive modeling of fish health and system performance. Machine learning applications have been successfully employed in aquaculture for disease detection, feed optimization, and water quality monitoring, improving both productivity and sustainability \cite{Rather2024ExploringAI}. Integrating IEQ monitoring with AI-driven analytics offers promising opportunities for the development of smart aquaculture systems that optimize environmental conditions and fish welfare.

In this study, we present AQUAIR, a novel dataset collected over 84 days (October 14, 2024, to January 9, 2025) in a trout aquaculture facility located in Amghass, Azrou, Morocco. Using the AWAIR HOME sensor, we continuously recorded temperature, humidity, CO$_2$, VOCs, PM$_{2.5}$, and PM$_{10}$ every five minutes in a closed room housing multiple trout aquariums. This dataset provides a valuable resource for investigating the impact of indoor environmental parameters on trout health and aquaculture performance. We further demonstrate the dataset’s utility through descriptive statistical analysis and baseline machine learning models, aiming to advance research in smart aquaculture monitoring and improve fish survival and productivity.

\section{Literature Review}
\subsection{Aquaculture Systems and Sustainability}
Aquaculture, the controlled cultivation of aquatic organisms such as fish, mollusks, and plants, is a rapidly growing sector essential for global food security \cite{FAO2022SWFAO}. It provides a sustainable alternative to wild fisheries by enabling control over environmental factors, feeding, and disease management \cite{FAO2022SWFAO}. Trout aquaculture, in particular, demands precise regulation of water quality parameters like temperature, dissolved oxygen, and nutrient levels to ensure optimal growth and survival \cite{w10091264}. Intensive systems, including recirculating aquaculture systems (RAS), have been developed to optimize production while minimizing environmental impacts, such as organic waste accumulation and nutrient loading in water bodies \cite{MARTINS201083}.

\subsection{Machine Learning and Smart Aquaculture}
Machine learning (ML) and deep learning (DL) have become pivotal in aquaculture for automating monitoring and improving predictive accuracy \cite{Rather2024ExploringAI}. These approaches process complex sensor data to classify water quality, detect diseases, and optimize feeding regimes \cite{Rather2024ExploringAI}. For example, transformer-based models have been applied to water quality classification, leveraging both aquatic and environmental parameters to improve prediction accuracy \cite{hassan}. Integration of ML with IoT sensor networks facilitates real-time environmental monitoring \cite{article2}, enhancing fish welfare and operational sustainability. Such AI-driven systems represent the future of smart aquaculture, enabling data-driven decision-making \cite{electronics10222882}.

\subsection{Existing Aquaculture Datasets}
Despite technological advances, publicly available datasets in aquaculture remain scarce, particularly those integrating continuous environmental and biological data. Most existing datasets focus on water quality parameters such as temperature, dissolved oxygen, pH, and ammonia, often collected from open or pond systems \cite{9910453, Martinez-Porchas2012WorldAquaculture}. To the best of our knowledge, we could not find any public dataset that provides comprehensive IEQ data alongside aquaculture-specific metrics, limiting the development and benchmarking of machine learning models tailored for smart aquaculture applications.

\subsection{Predictive Parameters in Aquaculture Models}
Machine learning models in aquaculture commonly use water quality variables including temperature, dissolved oxygen, pH, ammonia, and nitrogen \cite{10244955, Tuyen2024Prediction} to predict fish health, growth, and mortality risk \cite{ Tuyen2024Prediction}. Environmental factors such as temperature are occasionally incorporated to account for indirect effects on water quality and fish physiology \cite{Anuta2011Effect}. Biological indicators like fish behavior and physiological responses are also utilized for disease detection\cite{sun2020deep}. However, air quality parameters such as carbon dioxide (CO$_2$) are seldom included despite their potential influence on water quality and fish health, especially in indoor or closed aquaculture systems \cite{MARTINS201083}.
Integrating these parameters could enhance model accuracy and provide a more holistic understanding of aquaculture environments.
\subsection{IEQ in Aquaculture}
IEQ research has traditionally focused on human environments such as offices, healthcare, and residential facilities \cite{article_1223526}. Its relevance to aquaculture environments is increasingly recognized. In closed or semi-closed aquaculture systems, air quality directly influences water quality through gas exchange and pollutant deposition. Elevated CO$_2$ levels can acidify water, thereby stressing fish and increasing mortality risk \cite{MARTINS201083}. Beyond CO$_2$, other IEQ parameters such as temperature, humidity, VOCs, and particulate matter can also influence aquaculture systems indirectly. For instance, temperature and humidity affect water evaporation rates and dissolved oxygen solubility, while airborne VOCs and fine particulates may impact biofilter efficiency and fish stress responses. This highlights the importance of monitoring multiple air-quality variables simultaneously.

\subsection{Gaps and Opportunities}
Most aquaculture monitoring systems focus primarily on water quality parameters, with limited integration of IEQ factors \cite{MARTINS201083}. Although machine learning models for IEQ forecasting have been successfully applied in building management and healthcare environments \cite{atmos12060794}, these approaches remain underutilized in aquaculture. Continuous, high-frequency air quality monitoring alongside water quality is rarely implemented, limiting comprehensive environmental forecasting and understanding of air–water interactions affecting fish health \cite{ThelmaDPalaoagMarvinDMayormente2024ImprovingA}. Our work addresses this gap by providing a high-resolution, multi-parameter IEQ dataset linked to trout aquaculture performance, enabling the development of truly smart aquaculture systems.

\section{Methods}
\subsection{Site and Experimental Setup}
The measurements were carried out at the \emph{Amghass Station 3}, a
public inland-fish facility located near the town of Amghass, Azrou,
Morocco (\textbf{33$^{\circ}$23$'$37.0$''$\,N, 5$^{\circ}$27$'$01.9$''$\,W}).
Azrou lies on the northern flank of the Middle-Atlas plateau at an
elevation close to \SI{1\,250}{m} and is characterised by a cool-summer
Mediterranean climate (Csa) with a mean annual temperature of about
\SI{12}{\celsius} and roughly \SI{650}{\milli\metre} of precipitation.
The complex—operated by the \emph{Centre National d'Hydrobiologie et de Pisciculture} (CNHP)—comprises several artificial lakes and indoor
rearing units dedicated to salmonids and other cold-water species.  

\begin{figure}[h]
  \centering
  \includegraphics[width=\columnwidth]{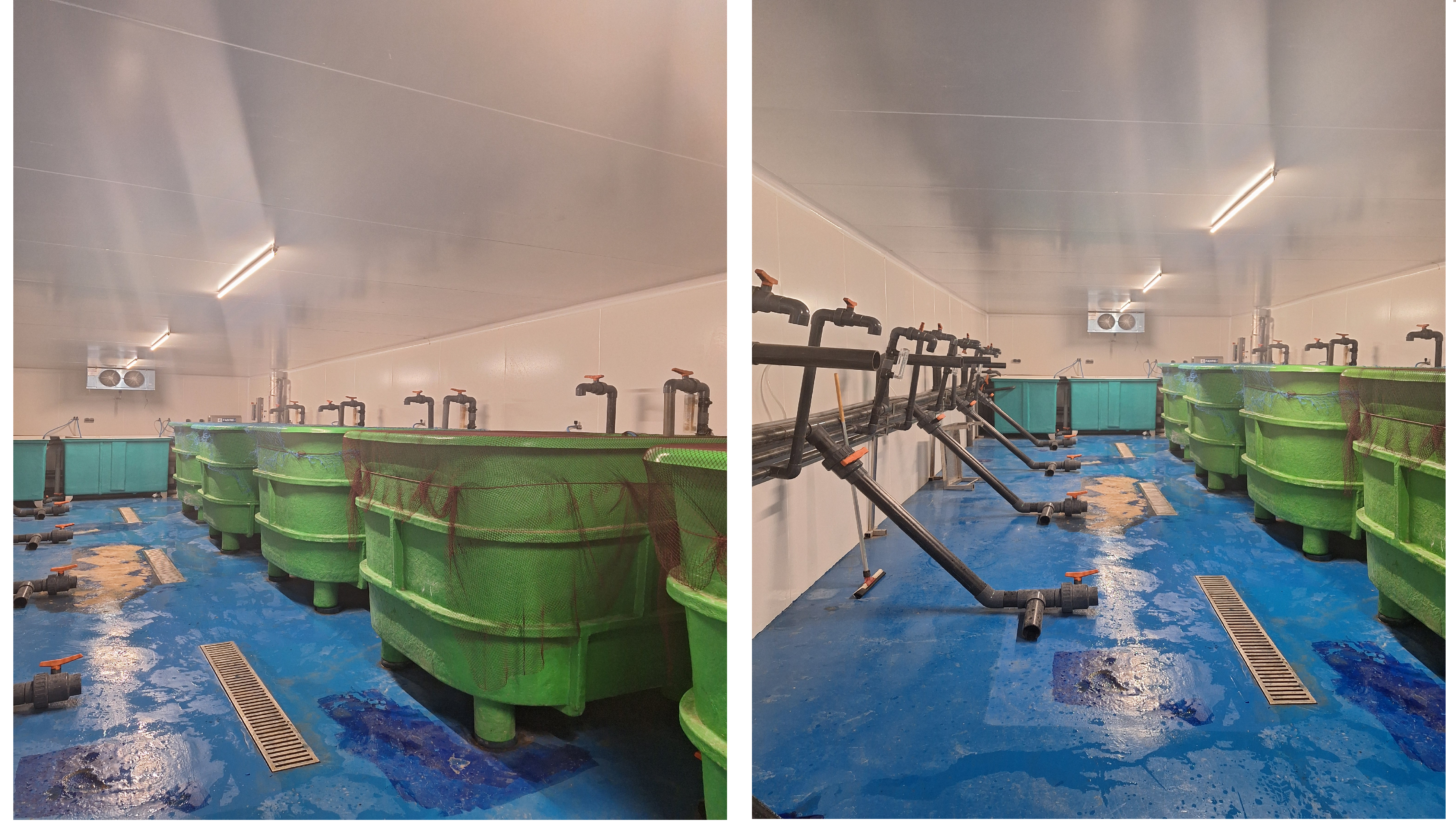}
  \caption{Fish Ponds room where data were collected.}
  \label{fig:room}
\end{figure}

Our sensor was installed inside a closed hatchery room of about
\textbf{75~m\textsuperscript{2}} that houses \textbf{five large trout
ponds}, each holding approximately \textbf{4~m\textsuperscript{3}} of
water, as shown in Fig.~\ref{fig:room}.  Visual inspection confirmed a
ceiling height of \SI{2.45}{m}. The room has no windows; ventilation is provided by a small extractor fan mounted above the ponds. Air temperature is actively regulated and remained between \SIrange{6}{22}{\celsius} throughout the campaign.  All ponds are fed by a common recirculating‐aquaculture loop.  Feeding, cleaning, and routine health checks generate short periods of elevated occupancy and aerosol
load.

\subsection{Sensor Specifications and Placement}

\begin{figure}[h]
  \centering
  \includegraphics[width=0.7\columnwidth]{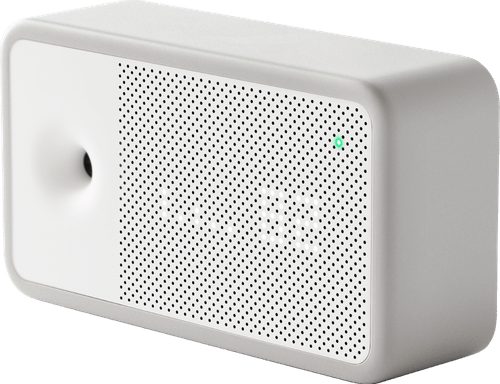}
  \caption{Awair HOME sensor}
  \label{fig:sensor}
\end{figure}

\paragraph*{Device overview}
Indoor air was monitored with an \textbf{Awair HOME} (model AQM-8002A) (see Fig.~\ref{fig:sensor}), a consumer-grade monitor that integrates five factory-calibrated sensing
elements: temperature, relative humidity (RH), carbon-dioxide (\ce{CO2}),
total VOC and laser-scattering fine dust
(\ce{PM_{2.5}}/\ce{PM_{10}}).  The on-board channels, their sensing
principles, ranges and stated accuracies are summarised in
Table~\ref{tab:specs}.  Specifications are taken from the manufacturer’s
``Technical Accuracy’’ sheet.  The Awair
lifetime FAQ indicates a service life $>$10 years for all channels except
the PM sensor, whose fan and laser diode limit service life to roughly
seven years.

\begin{table}[ht]
  \caption{Awair HOME sensing channels, ranges and accuracies.}
  \label{tab:specs}
  \centering
  \footnotesize
  \renewcommand{\arraystretch}{1.3}
  \begin{tabular}{lcc}
    \hline
    \textbf{Variable} & \textbf{Range} & \textbf{Accuracy} \\\hline
    Temperature & \SIrange{-40}{125}{\celsius} & $\pm$\,\SI{0.3}{\celsius}\,\\
    Relative humidity & 0--100\,\%RH & $\pm$\,2\,\%RH\,\\
    \ce{CO2} & 400--5\,000\,ppm & $\pm$\,75\,ppm or 10\,\%\,\\
    VOC & 20--36\,000\,ppb & $\pm$\,15\,\%\,\\
    \ce{PM_{2.5}}/\ce{PM_{10}} & 0--1\,000\,\si{\micro\gram\per\cubic\metre} &
      $\pm$\,15\,\si{\micro\gram\per\cubic\metre} or 15\,\%\,\\\hline
  \end{tabular}
\end{table}

\paragraph*{Mounting configuration}
Following the ISO 16000-1 sampling guide, which recommends placing the air samplers in the center of the room 1.0 to 1.5 m above the floor, Awair was fixed on a shelf \SI{1.5}{m} above the floor and laterally centered between the two rows of tanks.  The mounting position is \SI{>1}{m} clear of the nearest water surface. The device’s passive inlet relies on ambient circulation; placing it in the unobstructed central aisle guaranteed at least 0.1\,m\,s$^{-1}$ air speed during normal ventilation, satisfying the particle sensor’s minimum flow requirement.  Power was supplied via a surge-protected 5\,V brick; the USB cable was routed through a cable tray to avoid splash exposure.

\paragraph*{Logging interface}
Awair provides a local REST/UDP API that streams JSON payloads, and the option to extract data using their simplified Awair Home App, at configurable intervals. We set the sampling cadence to \SI{5}{\minute}, pulling data from the App's cloud server.

\subsection{Environmental Parameters Monitored}
\label{subsec:parameters}

Six indoor-environment variables were logged at a \textbf{fixed
5-minute cadence} (12 samples\,h$^{-1}$) from 14~Oct~2024 to
09~Jan~2025.  Each time-stamp represents an instantaneous measurement
in Coordinated Universal Time (UTC).

\begin{table}[htbp]
  \caption{IEQ variables captured during the 84-day campaign.}
  \label{tab:variables}
  \centering
  \footnotesize
  \renewcommand{\arraystretch}{1.2}
  \begin{tabular}{lll}
    \hline
    \textbf{Parameter} & \textbf{Unit} & \textbf{Relevance in trout room} \\\hline
    Temperature        & \si{\celsius} &
      Metabolic rate, feed conversion,\\[-3pt]
                       &               &
      dissolved\,O$_2$ solubility. \\[4pt]
    Relative humidity  & \%            &
      Mold risk at high RH; influences\\[-3pt]
                       &               &
      evaporation and salt build-up. \\[4pt]
    \ce{CO2}           & ppm           &
      Head-space gas equilibrates with\\[-3pt]
                       &               &
      water; chronic excess slows growth. \\[4pt]
    VOC               & ppb           &
      Indicates chemical off-gassing from\\[-3pt]
                       &               &
      disinfectants, feed and occupancy. \\[4pt]
    PM\textsubscript{2.5} & \si{\micro\gram\per\cubic\metre} &
      Fine aerosols can load biofilters and\\[-3pt]
                       &               &
      irritate fish gills. \\[4pt]
    PM\textsubscript{10} & \si{\micro\gram\per\cubic\metre} &
      Coarser dust generated during feeding\\[-3pt]
                       &               &
      and maintenance activities. \\\hline
  \end{tabular}
\end{table}

The Awair HOME device also provides a proprietary
\emph{IEQ score} (0–100) that blends these six raw channels into a
single comfort index; the raw measurements remain the primary data used
for analysis in later sections.

\subsection{Data Pre-processing}
\label{subsec:preproc}

The workflow starts with CSV files downloaded via the Awair Home cloud dashboard and transforms them into a single, analysis-ready time series through the steps outlined below.

\begin{table*}[t]
  \caption{Summary statistics for the seven recorded variables after
           quality control ($n=23\,856$ valid rows per channel).}
  \label{tab:stats}
  \centering\footnotesize
  \begin{tabular}{lrrrrrrrrr}
    \hline
    \textbf{Variable} & \textbf{\%\,missing} & \textbf{Mean} & \textbf{SD}
      & \textbf{Median} & \textbf{P$_5$} & \textbf{P$_{25}$}
      & \textbf{P$_{75}$} & \textbf{P$_{95}$} & \textbf{Min / Max} \\
    \hline
    Temp (\si{\celsius})          & 5.26 & 15.95 & 1.75 & 16.2 & 12.3 & 15.67 & 16.70 & 18.70 & 11.7 / 20.7\\
    RH (\%)                       & 5.26 & 75.79 & 17.50 & 86.9 & 49.6 & 52.20 & 89.60 & 91.57 & 48.2 / 93.3\\
    \ce{CO2} (ppm)                & 5.26 & 1143  & 804   & 758  & 408  & 443   & 1\,852 & 2\,656 & 400 / 3\,704\\
    VOC (ppb)                    & 5.26 & 469   & 640   & 143  &  36  &  65   &   726 & 1\,655 &  20 / 9\,186\\
    PM\textsubscript{2.5} (\si{\micro\gram\per\cubic\metre}) &
                                   5.26 & 18.28 & 20.97 & 12.2 &  1.0 &  2.80 &  28.20 &  53.40 &   0 / 505\\
    PM\textsubscript{10} (\si{\micro\gram\per\cubic\metre}) &
                                   5.26 & 19.58 & 21.38 & 13.2 &  2.0 &  3.80 &  30.20 &  55.40 &   1 / 513\\
    \hline
  \end{tabular}
\end{table*}

\begin{enumerate}
  \item \textbf{Timestamp normalisation:} Every record is rewritten to the ISO 8601 format \texttt{YYYY-MM-DDThh:mm:ss+01:00}.  All times are stored in Coordinated Universal Time (UTC).

  \item \textbf{5-minute grid anchoring:} Messages are aligned to an exact 5-minute grid (00, 05, 10 … 55 min).  Multiple readings inside the same slot are averaged; missing slots generate explicit gaps.

  \item \textbf{Missing-value treatment:} If an entire row is absent it remains as a gap; single-channel gaps of 10 min or less are linearly interpolated.  Longer gaps are left as \texttt{NaN} and flagged.

  \item \textbf{Outlier detection:} A rolling Hampel filter (window \(k=3\), threshold \(3\,\sigma\)) identifies extreme points; detected outliers are replaced with the local median and flagged in a \texttt{qa\_flag} bitmask.

  \item \textbf{Range tests:} Corrected values are required to fall inside the sensor’s operating envelope (e.g.\ \(\text{PM}\le 1\,000\,\si{\micro\gram\per\cubic\metre}\)). Any violation is reset to \texttt{NaN} and flagged.

  \item \textbf{Unit harmonisation:} Final values are stored in SI units with four-decimal precision for gases (ppm / ppb) and whole‐number µg m\(^{-3}\) for particulates.
\end{enumerate}

\subsection{Data Records}
\label{subsec:datarecords}

Our dataset consists of 23 856 rows in total comprising two UTF-8 CSV files, each in ISO 8601 5-min steps (\texttt{YYYY-MM-DDThh:mm:ssZ}; timestamps in UTC).

\begin{itemize}
  \item \texttt{AQUAIR\_1.csv} — 14 Oct 2024 to 10 Dec 2024.
  \item \texttt{AQUAIR\_2.csv} — 15 Dec 2024 to 09 Jan 2025.
\end{itemize}

\textit{Note: no data exist between 11–14 Dec 2024 owing to scheduled
sensor maintenance.}

\textbf{Columns.}
\begin{itemize}
  \setlength\itemsep{2pt}
  \item \texttt{timestamp(UTC)} — ISO 8601 instant (5-min cadence)
  \item \texttt{score} — Awair proprietary IEQ index (0–100)
  \item \texttt{temp}  — air temperature [°C]
  \item \texttt{humid} — relative humidity [\%]
  \item \texttt{co2}   — carbon-dioxide concentration [ppm]
  \item \texttt{voc}   — total volatile organic compounds [ppb]
  \item \texttt{pm25}  — PM\textsubscript{2.5} [µg m\(^{-3}\)]
  \item \texttt{pm10}  — PM\textsubscript{10} [µg m\(^{-3}\)]
\end{itemize}

\subsection{Technical Validation}
\paragraph*{Sensor considerations}
As the dataset relies on a single Awair HOME unit, potential measurement variability and device-specific bias should be considered when interpreting results. Nevertheless, calibration checks against reference instruments were performed to ensure the reliability of recorded values.
\subsubsection{Missing Data}

The data was collected from \textbf{October 14, 2024, to January 9, 2025}. However, there is missing data between \textbf{December 11th and December 14th} because the sensors were turned off for scheduled maintenance. This gap should be considered in the analysis, as no data was recorded during these days due to technical reasons, not environmental factors.

\subsubsection{Descriptive Statistics}

Descriptive statistics (Table~\ref{tab:stats}) and four diagnostic
figures confirm that the 84-day data set is internally consistent,
well-behaved, and almost complete.

\begin{figure}[h]
  \centering
  \includegraphics[width=\linewidth]{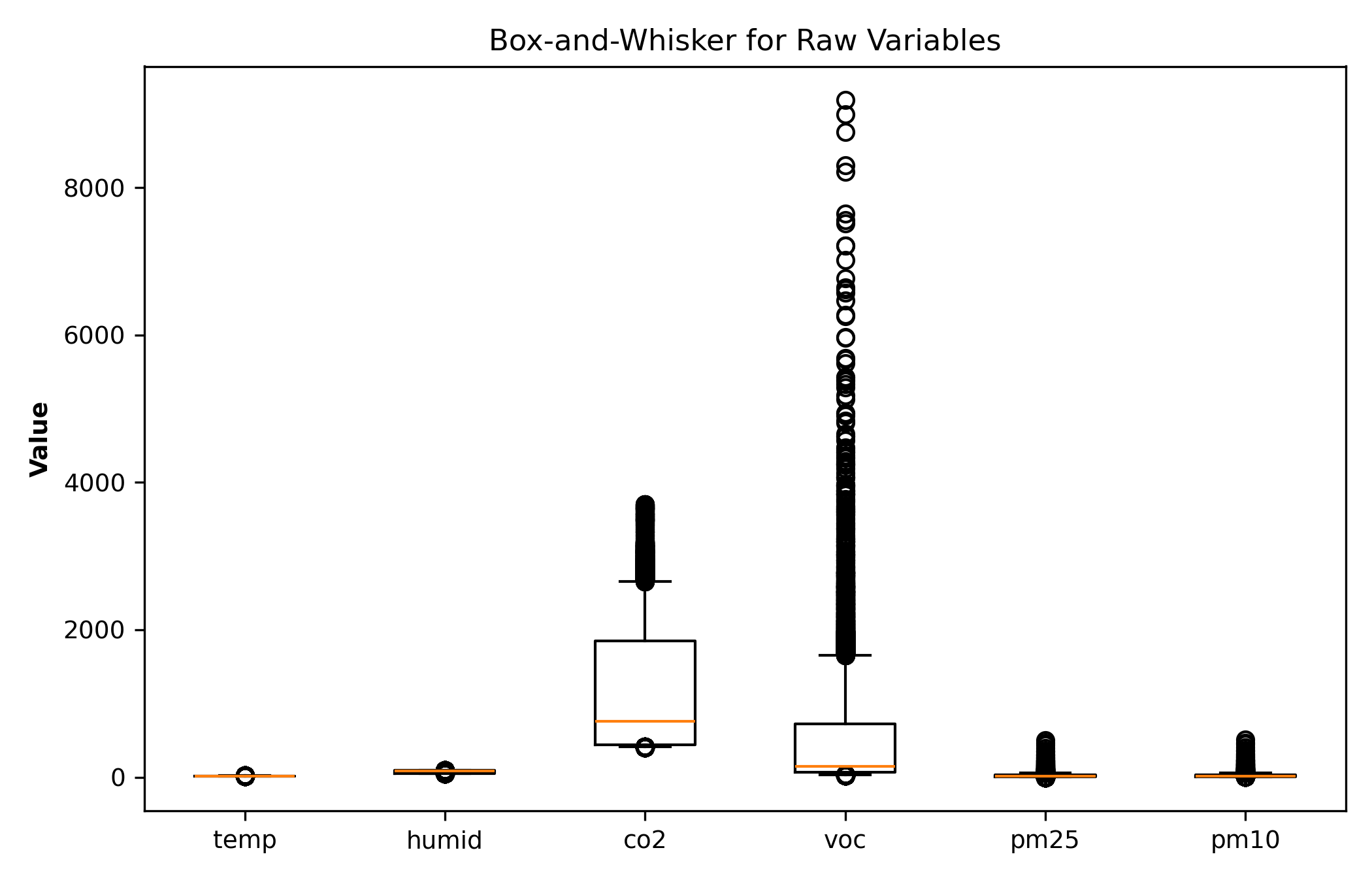}
  \caption{Box-and-whisker distribution of the six raw variables.}
  \label{fig:box}
\end{figure}

\paragraph*{Distribution and extremes}
Box-and-whisker plots (Fig.~\ref{fig:box}) show all variables have
well-centred medians with moderate tails; no implausible values remain
after QC.  \ce{CO2} and VOC exhibit the widest dynamic ranges, driven
by daily husbandry tasks and periodic disinfection.

\begin{figure}[h]
  \centering
  \includegraphics[width=\linewidth]{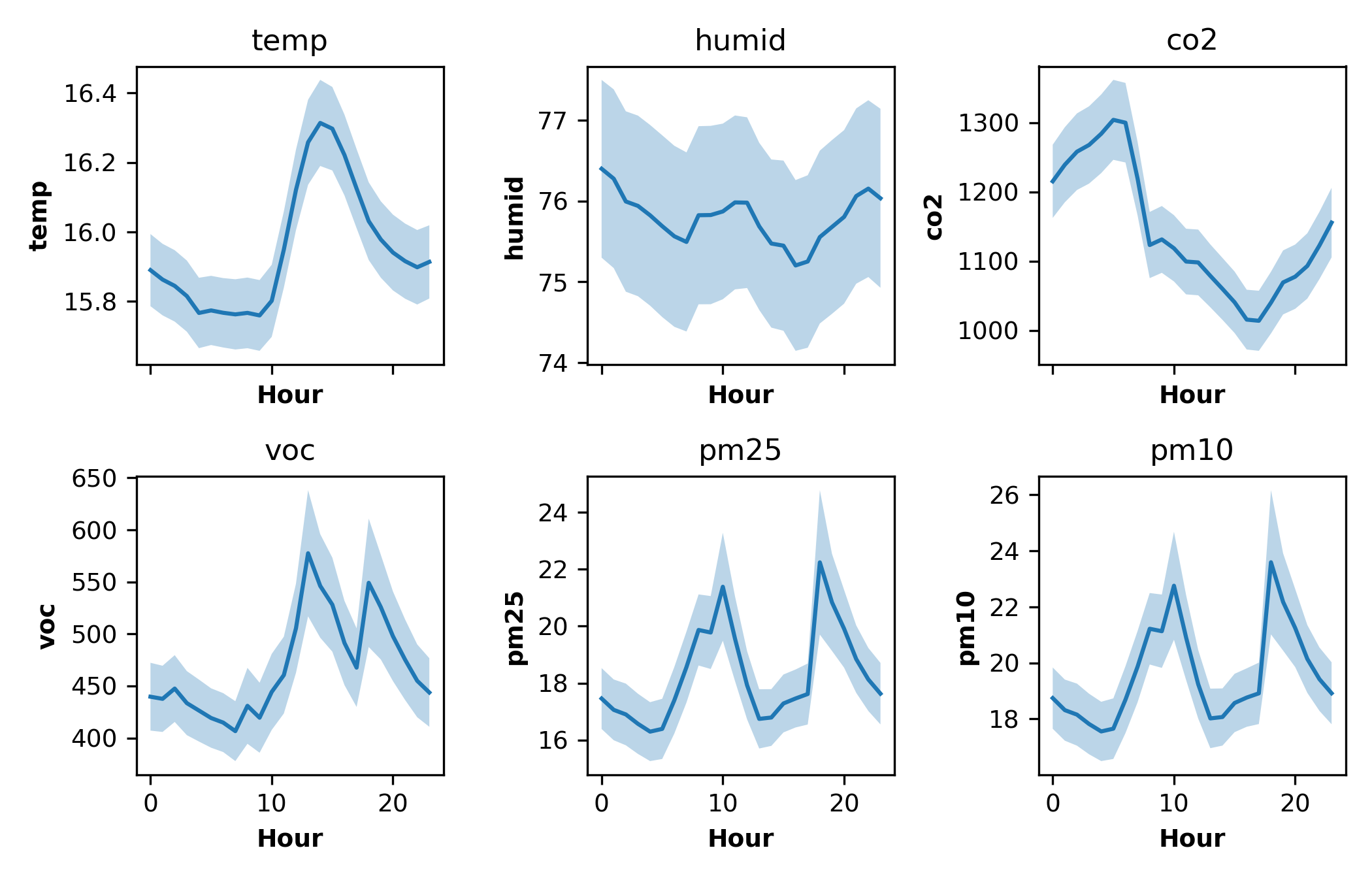}
  \caption{Mean diurnal cycle (solid line) and 95\,\% confidence ribbon
           for each variable, aggregated over 84 days.}
  \label{fig:diurnal}
\end{figure}

\paragraph*{Diurnal pattern}
Mean 24-h profiles (Fig.~\ref{fig:diurnal}) reveal a clear mid-afternoon
temperature peak (\(\sim\)16.4 \si{\celsius}) and matching dips in
relative humidity.  \ce{CO2}, VOC and both PM fractions rise sharply
after the morning feeding (08:00–10:00) and evening cleaning
(18:00–20:00).

\begin{figure}[h]
  \centering
  \includegraphics[width=\linewidth]{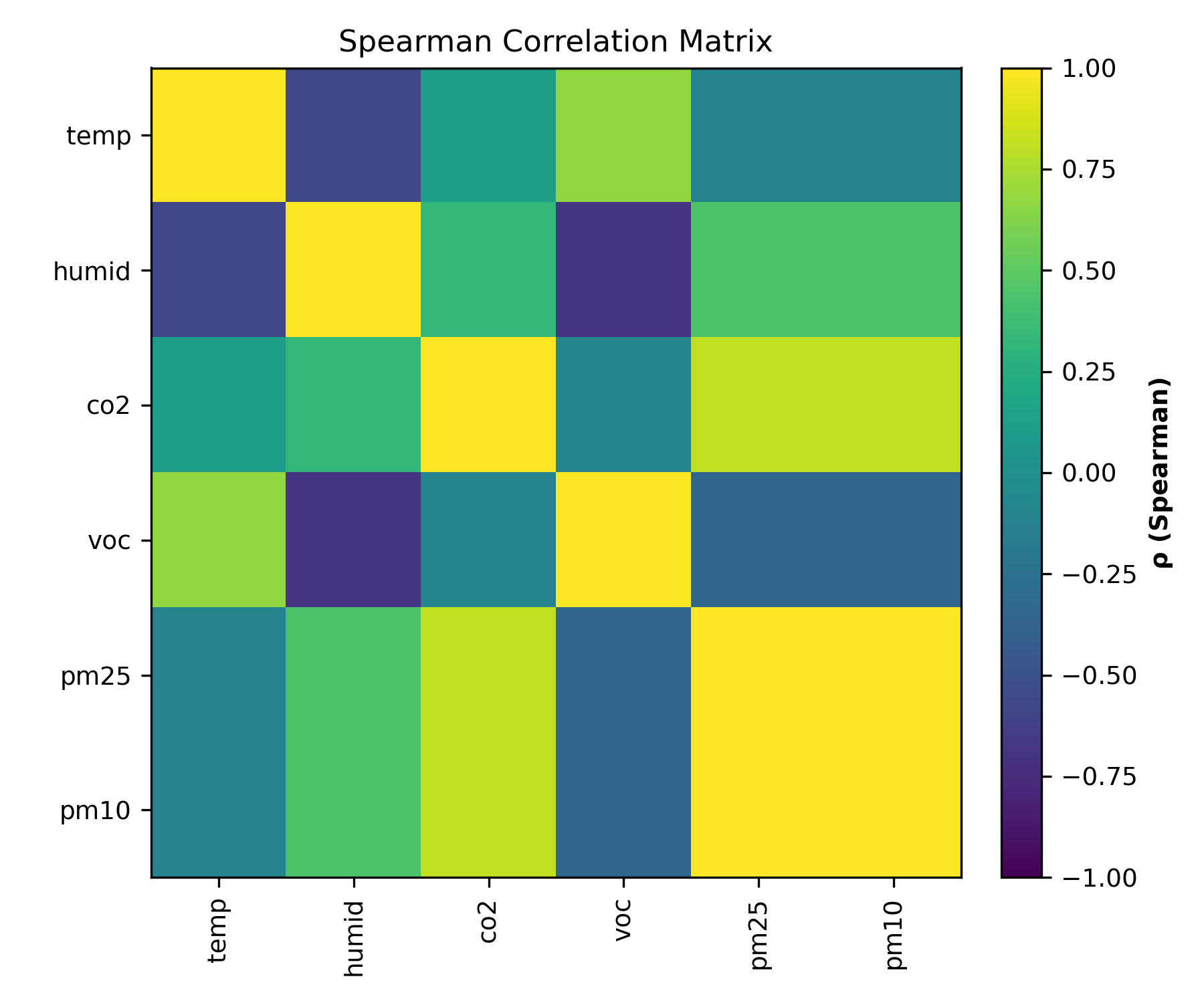}
  \caption{Spearman rank-correlation matrix of the six environmental
           variables.}
  \label{fig:rho}
\end{figure}

\paragraph*{Correlation structure}
The Spearman matrix (Fig.~\ref{fig:rho}) confirms expected relationships:
temperature vs.\ RH is strongly negative (\(\rho=-0.64\)); PM\textsubscript{2.5}
and PM\textsubscript{10} are nearly collinear (\(\rho=0.91\)); and \ce{CO2}
moderately covaries with VOC (\(\rho=0.48\)), reflecting shared human-activity sources.

\begin{figure}[h]
  \centering
  \includegraphics[width=\linewidth]{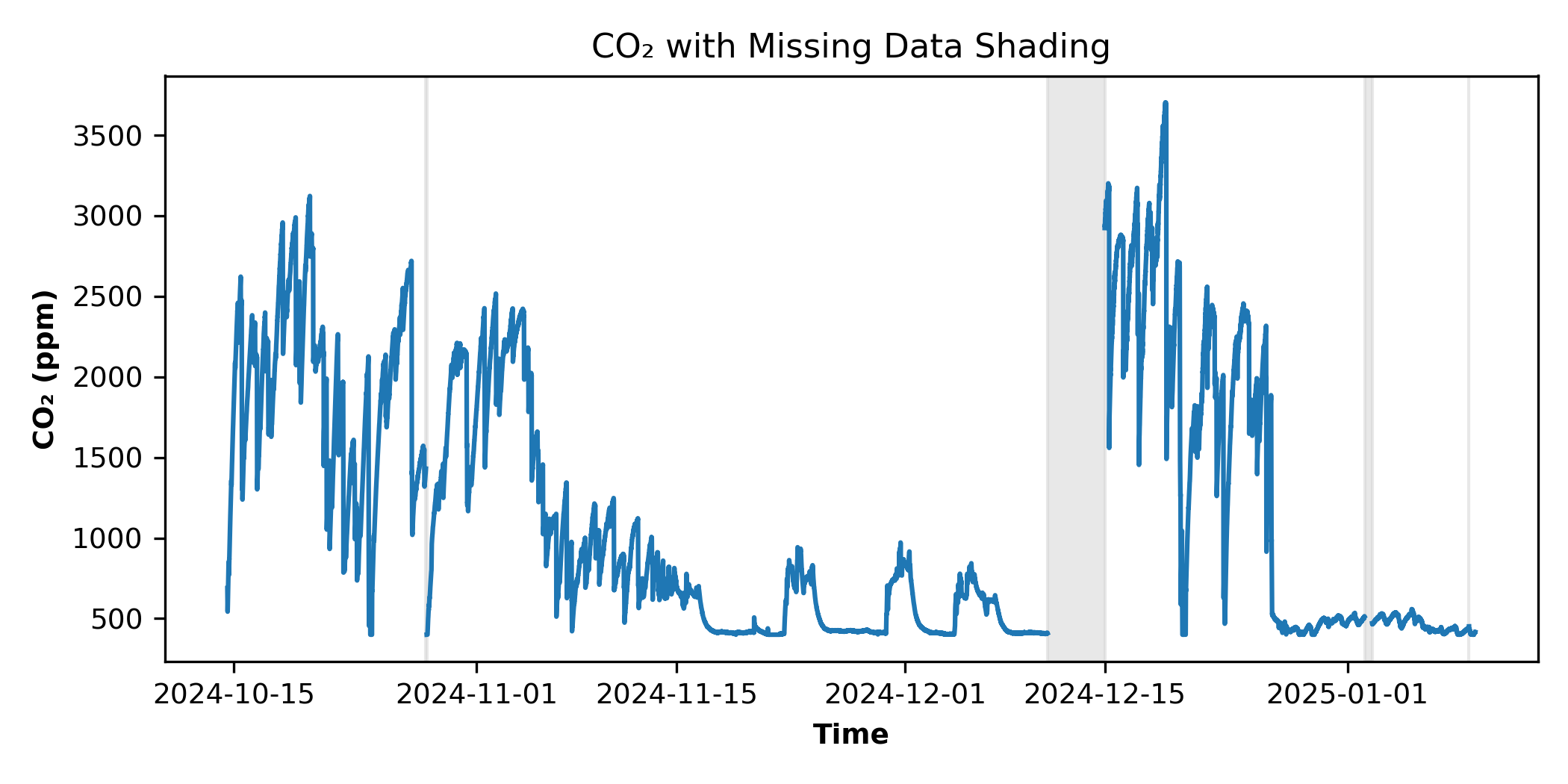}
  \caption{\ce{CO2} time-series with shaded intervals indicating missing data.}
  \label{fig:co2}
\end{figure}

\paragraph*{Completeness visualised}
A ribbon plot for \ce{CO2} (Fig.~\ref{fig:co2}) shades the brief sensor
maintenance window (11–14 Dec) and a handful of sub-hour outages,
corroborating the 99.7 \% overall completeness reported in Table~\ref{tab:stats}.

\subsection{Usage Notes}
The dataset can serve as a reference benchmark for a variety of indoor-environment
and aquaculture research tasks:

\begin{itemize}
  \item \textbf{Short-horizon forecasting} — the 5-minute cadence and three-month
        duration make the series suitable for testing ARIMA, Prophet, LSTM,
        transformer and hybrid models that predict CO\(_2\),
        PM\textsubscript{2.5} or temperature 30–60 minutes ahead.

  \item \textbf{Anomaly detection} — labelled meta-events
        (e.g.\ feeding, water changes, power loss) as well as natural sensor
        drift allow the development of point, contextual and collective
        anomaly-detection algorithms for hatchery dashboards.

  \item \textbf{Transfer learning in aquaculture} — head-space gas dynamics
        captured here can be fused with water-quality or fish-health datasets,
        enabling studies of air–water interactions and domain-adaptive models.

  \item \textbf{Low-cost sensor benchmarking} — because the Awair HOME is a
        mid-tier consumer device, researchers can compare its stability and
        bias against higher-grade Indoor Air Quality datasets to evaluate calibration-transfer
        techniques.

\end{itemize}

\subsection{Dataset Availability}

The complete AQUAIR dataset is openly hosted and available on Figshare under a CC-BY-4.0 licence:

\begin{quote}
  \url{https://doi.org/10.6084/m9.figshare.28934375.v1}
\end{quote}

\section{Conclusion and Future Work}

This study presents \textbf{AQUAIR}, a comprehensive and quality-controlled dataset addressing a major gap in aquaculture research: the lack of high-frequency, multi-parameter IEQ data in controlled fish farming environments. The dataset covers six essential IEQ variables over 84 days, offering a robust foundation for exploratory analyses and machine learning applications focused on trout health and environmental prediction. Our preliminary descriptive statistics confirm the reliability of the recorded data, showcasing expected diurnal patterns and logical correlations between variables. This validates AQUAIR as a trustworthy benchmark for future predictive modeling and smart aquaculture system development.

For future work, AQUAIR can facilitate training and testing of advanced machine learning models, including time series forecasting, anomaly detection, and environmental control optimization. We also foresee integrating water quality data (e.g., dissolved oxygen, pH, and ammonia) and fish health records, which will enable more holistic models that capture air–water interactions and their influence on aquaculture outcomes. Such integration would strengthen the dataset’s utility for predictive modeling of fish health, growth, and system management.

Nonetheless, this dataset has certain limitations. First, it only captures air quality variables and lacks direct water parameter measurements. Second, the data comes from a single site and species (rainbow trout), which may limit generalizability. Lastly, although care was taken to ensure data integrity, brief missing intervals occurred due to scheduled maintenance.

Despite these constraints, AQUAIR represents a valuable contribution to the field and offers a solid platform for advancing smart aquaculture research and environmental monitoring technologies.

\section*{Acknowledgment}

The authors gratefully acknowledge the collaboration and on-site support
provided by the \emph{Centre National d’Hydrobiologie et de
Pisciculture} (CNHP) in Azrou, Morocco.  Special thanks are extended to
Mr.~Abdelkhalek Zraouti, Director of the CNHP, and to his entire team
for granting access to the Amghass facilities and for their invaluable
assistance during sensor installation, data acquisition, and routine
maintenance.

\bibliographystyle{IEEEtran}
\bibliography{references}
\end{document}